\documentclass[a4paper,10pt,twoside]{article}

\usepackage{clin}        
\usepackage{harvard}     

\usepackage{tikz}
\usetikzlibrary{shapes,arrows,positioning,calc}
\usepackage{paralist}
\usepackage{tablefootnote}
\usepackage{caption}
\usepackage{subfigure}
\usepackage{gb4e}

\begin{document}

\title{MoNoise: Modeling Noise Using a Modular Normalization System.}

\author{Rob van der Goot \email{r.van.der.goot@rug.nl}\\
{\normalsize \bf Gertjan van Noord} \email{g.j.m.van.noord@rug.nl}
\AND \addr{University of Groningen, Broerstraat 5, 9712CP, Groningen, Netherlands}}

\maketitle\thispagestyle{empty} 


\begin{abstract}
We propose MoNoise: a normalization model focused on generalizability and
efficiency, it aims at being easily reusable and adaptable. Normalization is
the task of translating texts from a non-canonical domain to a more canonical
domain, in our case: from social media data to standard language. Our proposed
model is based on a modular candidate generation in which each module is
responsible for a different type of normalization action. The most important
generation modules are a spelling correction system and a word embeddings
module. Depending on the definition of the normalization task, a static lookup
list can be crucial for performance. We train a random forest classifier to
rank the candidates, which generalizes well to all different types of
normalization actions. Most features for the ranking originate from the
generation modules; besides these features, N-gram features prove to be an
important source of information. We show that MoNoise beats the
state-of-the-art on different normalization benchmarks for English and Dutch,
which all define the task of normalization slightly different.

\end{abstract}

\section{Introduction}
\label{sec:introduction}
The spontaneous and diverse nature of language use on social media leads to
many problems for existing natural language processing models. Most
existing models are developed with a focus on more canonical language. These
models do not cope with the disfluencies and unknown phenomena occurring in
social media data. This is also known as the problem of domain adaptation: in
which we try to adapt a model trained on a \texttt{source} domain to another
\texttt{target} domain.  Solutions for this problem can be broadly divided in
two strategies: adapting the model to the target domain, or adapting the data
to the source domain~\cite{eisenstein:2013:NAACL-HLT}.

Domain adaptation by adapting the model to the target domain can be done in
different ways. The most straightforward method is to train the model on
annotated data from the target domain. Newly annotated data can be obtained by
hiring human annotators. However, it is cheaper to annotate data automatically
using an existing model. This is called self-training, or up-training if
the data is annotated by an external model. The effect of adding this newly
annotated data depends on the nature of the new data compared to the data of
the target and source domain. The added data should be annotated with a high
accuracy, so it can not be too distant from the source domain. However, it
should add some information inherent to the target domain. There is ample
previous work in this direction in which different strategies of up-training
are
used~\cite{foster2011hardtoparse,khan-dickinson-kubler:2013:RANLP-2013,petrov2012overview}.

The other strategy for domain adaptation is to convert the data to the source
domain; this is the strategy explored in this work. This task is often referred
to as normalization, because we aim to convert data from the target domain to the
more 'normal' source domain, for which a model is already available. The main
advantage of this approach is that we only need one normalization model, which
we can use as preprocessing step for multiple natural language processing systems.

Normalization is a subjective task; the goal is to convert to `normal'
language. At the same time, we must preserve the meaning of the original
utterance. This task comprises the correction of unintentional anomalies (spell
correction) as well as intentional anomalies (domain specific language
phenomena).  Annotator disagreement can thus have two sources, the decision
whether a word should be normalized, and the choice of the correct
normalization candidate.  We discuss these problems in more depth in
Section~\ref{sec:dataNorm}. In the rest of this paper we will use the term 
`anomaly' for words in need of normalization according to the annotators.

\begin{exe}
    \ex
    \gll Ima regret bein up this late 2mr \\
    I'm going to regret being up this late tomorrow  \\
    \label{Tweet1}
\end{exe}

Example \ref{Tweet1} shows that the normalization task comprises of
different types of transformations. Replacements like `bein' $\mapsto$ `being' 
are quite similar on the surface, whereas `tmr' $\mapsto$ tomorrow shows that
we need more than edit distances on the character level. This example 
includes a 1-N replacement (Ima $\mapsto$ `I'm going to', meaning that a single
token is mapped to multiple tokens. Not all the annotated corpora we use
include annotation for these cases (see Section \ref{sec:dataNorm}).
1-N replacements show a strong Zipfian distribution in the corpora that include
them in the annotation, because of the expansion of phrasal abbreviations like
`lol' and `idk' which are very common. Some of the corpora also include N-1
replacements, meaning the merging of two consecutive words; however, this is a
very rare phenomenon.

Because the normalization problem comprises of a variety of different
normalization actions required for different types of anomalies, we propose
to tackle this problem in a modular way.  Different modules can then be
designed for different types of anomalies.  Our most important modules are:
a spell correction module, a word embeddings module, and a static lookup list
generated from the training data.  We use features from the generation modules
as well as additional features in a random forest classifier, which decides
which candidate is the correct normalization. We experiment with a variety of
additional features, of which the N-gram features are by far the best
predictor.

The rest of this paper is structured as follows: We first discuss related work
(Section~\ref{sec:related}), after which we shortly describe the used data
(Section~\ref{sec:data}).  Next follows the methodology section
(\ref{sec:method}) and the evaluation section (\ref{sec:evaluation}), which are
both splitted by the two different parts of our system; candidate generation
and candidate ranking.  Finally, we conclude in Section~\ref{sec:conclusion}.

\section{Related Work}
\label{sec:related}
The first attempts at normalizing user generated content were focused on SMS
data;~\citeasnoun{choudhury2007investigation} annotated a dataset for this
domain, and reported the first results. They use a Hidden Markov Model
encoding based on characters and phonemic transcriptions to model the word
variation in SMS data.  The Viterbi algorithm is then used to find the most
likely replacement for each position.

Later, focus shifted towards normalization for social media, more specifically:
the Twitter domain. The first work on normalization for this domain was from
\citeasnoun{han-baldwin:2011:ACL-HLT2011}. They released the LexNorm corpus,
consisting of 549 Tweets annotated with their normalization on the word level.
Annotation is restricted to word to word replacements, so words like `gonna'
are kept untouched. \citeasnoun{han-baldwin:2011:ACL-HLT2011} also reported the
first results on this dataset.  They train a support vector machine which
predicts if a word needs normalization based on dependency tree context; the
length of the arcs and the head words are used as predictors. After this, they
generate candidates using a combination of lexical and phonetic edit distances.
Candidates are ranked using a combination of dictionary lookup, word similarity
and N-gram probabilities. Note that on this corpus, gold error detection is
usually assumed, accuracy is reported on only the words that need
normalization.

Over the years, many different approaches have been benchmarked on this
dataset; ~\citeasnoun{li-liu:2012:PAPERS} experiment with character based machine
translation.  \citeasnoun{hassan-menezes:2013:ACL2013} use random walks in a
bipartite graph based on words and their contexts to generate normalization
candidates, which they rank using the Viterbi algorithm. A log-linear model was
explored by~\citeasnoun{yang-eisenstein:2013:EMNLP}, they use sequential Monte
Carlo to approximate feature expectation, and rank them using a
Viterbi-encoding. Whereas most previous work normalizes on the word level or
character level, ~\citeasnoun{xu-xia-lee:2015:ACL-IJCNLP} attempt to normalize
on the syllable level. They translate noisy words to sequences of syllables,
which can then be normalized to canonical syllables, which can in turn be
merged back to form normalization candidates.

To the best of our knowledge, \citeasnoun{li2015joint} reported the highest
accuracy on the LexNorm dataset. They rerank the results of six different
normalization systems, including machine translation systems, a character
sequence labeling model and a spell checker. Each normalization system suggests
one candidate.  A Viterbi decoding based on the candidates and their possible
POS tags is then used to rank the candidates. This joint approach is beneficial
for both tasks.

More recently, the 2015 Workshop on Noisy User-generated Text hosted a shared
task on lexical normalization~\cite{baldwin-EtAl:2015:WNUT}. They defined the
task slightly different compared to the annotation of the LexNorm corpus.
Annotation included 1-N and N-1 replacements. N-1 replacements indicate
merging, which occurs very rarely. For the shared task, gold error detection
was not assumed, and was part of the task. A total of 10 teams participated in
this shared task, using a wide variety of approaches. For generation of
candidates the most commonly used methods include: character N-grams,
edit-distances and lookup lists. Ranking was most often done by conditional
random fields, recurrent neural networks or the Viterbi algorithm.

The best results on this new benchmark were obtained
by~\citeasnoun{jin:2015:WNUT}. This model generates candidates based on a
lookup list, all possible splits and a novel similarity index: the Jaccard
index based on character N-grams and character skip-grams. This novel
similarity index is used to find the most similar candidates from a dictionary
compiled from the golden training data. ~\citeasnoun{jin:2015:WNUT} also tests
if it is beneficial to find similar candidates in the Aspell
dictionary\footnote{\url{www.aspell.net}}, but concludes that this leads to
over-normalization.  The candidates are ranked using a random forest
classifier, using a variety of features: frequency counts in training data, a
novel similarity index and POS tagging confidence. 

Most of the previous work has been on the English language, although there has
been some work on other languages. We will consider normalization for Dutch to
test if our proposed model can be effective for other languages.  There has
already been some previous work on normalization for the Dutch language.
~\citeasnoun{DECLERCQ14} annotated a normalization corpus consisting of three
user generated domains. They experiment on this data with machine translation
on the word and character level, and report a 20\% gain in BLEU score,
including tokenization corrections.  Building on this work,
~\citeasnoun{Schulz:2016:MTN:2906145.2850422} built a multi-modular model, in
which each module accounts for different normalization problems, including:
machine translation modules, a lookup list and a spell checker. They also
report improved results for extrinsic evaluations on three tasks: POS tagging,
lemmatization and named entity recognition.

Our proposed system is the most similar to~\citeasnoun{jin:2015:WNUT}; however,
there are many differences. The main differences are that we use
word-embeddings for generation and include N-grams features for ranking, which
can easily be obtained from raw text. This makes the system more general and
easier adaptable to new data. The system from~\citeasnoun{jin:2015:WNUT} is
more focused towards the given corpus, and might have more difficulties on data
from another time span or another social media domain.

\section{Data}
\label{sec:data}
The data we use can be divided in two parts: data annotated for normalization, and 
other data.

\subsection{Normalization Corpora}
\label{sec:dataNorm}
The normalization task can be seen as a rather subjective task; the annotators
are asked to convert noisy texts to `normal' language. The annotation
guidelines are usually quite
limited~\cite{orpheeGuidelines,lexnorm2015guidelines}, leaving space for
interpretation; which might lead to inconsistent annotation.
\citeasnoun{Pennell2014256} report a Fleiss’ $\kappa$ of 0.891 on the detection
of words in the need of normalization. They also shared the annotation efforts
of each annotator, we used this data to calculate the pairwise human
performance on the choice of the correct normalization candidate. This revealed
that the annotators agree on the choice of the normalized word in 98.73\% of
the cases.  Note that this percentage is calculated assuming gold error
detection.  \citeasnoun{baldwin-EtAl:2015:WNUT} report a Cohen's $\kappa$ of
0.5854 on the complete normalization task, a lot lower compared
to~\citeasnoun{Pennell2014256}. Hence, we can conclude that the inter-annotator
agreement is quite dependent on the annotation guidelines.  After the decision
whether to normalize, annotator agreement is quite high on the choice of the
correct candidate.

\begin{table}
    \centering
    \begin{tabular}{l l r r r r r} 
        Corpus            & Source                                    & Words     & Lang. & Caps & Multiword & \%normalized \\ 
        LexNorm1.2         & \citeasnoun{yang-eisenstein:2013:EMNLP} & 10,564     & en    & no   & no        & 11.6  \\
        LiLiu           & \citeasnoun{li-liu:2014:P14-3}            & 40,560     & en    & some & no        & 10.5  \\
        LexNorm2015     & \citeasnoun{baldwin-EtAl:2015:WNUT}       & 44,385     & en    & no   & yes       & 8.9   \\
        GhentNorm       & \citeasnoun{DECLERCQ14}               & 12,901     & nl    & yes  & yes       & 4.8   \\
    \end{tabular}
    \caption{Comparison of the different corpora used in this work.}
    \label{tab:corpora}
\end{table}

The main differences between the different normalization corpora are shown in
Table~\ref{tab:corpora}.  Note that we use the LexNorm1.2 corpus, which contains
some annotation improvements compared to the original LexNorm corpus.
The GhentNorm corpus is the only corpus fully
annotated with capitals, even though the capital-use is not corrected; it is
preserved from the original utterance~\cite{DECLERCQ14}. The multiword
column represents whether 1-N and N-1 replacements are included in the
annotation guidelines; the corpora which do include this, also include
expansions of commonly used phrasal abbreviations as `lol' and `lmao', and N-1
replacements are extremely rare.  There is some difference in the percentage of
tokens that are normalized, probably due to differences in filtering and
annotation.

To give a better idea of the nature of the data and annotation, we will 
discuss some example sentences below.

\begin{exe}
    \ex 
    \gll lol or it could b sumthn else ... \\
    lol or it could be something else ... \\
    \label{ex:chenli1}
\end{exe}

Example~\ref{ex:chenli1} comes from the LiLiu corpus, this example contains two
replacements. The replacements are subsequent words, which is not uncommon;
this leads to problems for using context directly. The replacement `b'
$\mapsto$ `be' is grammatically close, whereas the replacement of `sumthn'
$\mapsto$ `something' is more distant, and would be harder to solve with
traditional spelling correction algorithms.

\begin{exe}
    \ex 
    \gll $<$USERNAME$>$ i aint messin with no1s wifey yo lol \\
    $<$USERNAME$>$ i ain't messing with {no one's} wifey you {laughing out loud} \\
    \label{ex:lexnorm20151}
\end{exe}

Example~\ref{ex:lexnorm20151} is taken from the LexNorm2015 corpus. This
annotation also include 1-N replacements; `no1s' and `lol' are expanded. the
word `no1s' is not only splitted, but also contains a substitution of of a
number to it's written form; two actions are necessary. In contrast to the
previous example, here the token `lol' is expanded; this is a matter of
differences in annotation guidelines.  The annotator decided to leave the word
`wifey' as is, whereas it could have been normalized to wife, this reflects the
suggested conservativity described in the annotation
guidelines~\cite{lexnorm2015guidelines}.

\begin{exe}
    \ex 
    \gll $<$USERNAME$>$ nee ! :-D kzal no es vriendelijk doen lol \\
     $<$USERNAME$>$ nee ! :-D {ik zal} nog eens vriendelijk doen {laughing out loud} \\
    \label{ex:ghentnorm}
\end{exe}

Example~\ref{ex:ghentnorm} comes from the GhentNorm corpus. The word `ik' (I)
is often abbreviated and merged with a verb in Dutch Tweets, leading to `kzal'
which is correctly splitted in the annotation to `ik zal' (I will). `no' is
probably a typographical mistake, whereas `es' is a shortening based on
pronunciation. Similar to the LexNorm 2015 annotation, the phrasal abbreviation
`lol' is expanded.



\subsection{Other Data}
\label{data:other}
In addition to the training data, we also use some external data for our
features. The Aspell dictionaries for Dutch and English are used as-is,
including the expansions of words\footnote{Obtained by using \texttt{-dump}}.
Furthermore, we use two large raw text databases; one with social media data,
and one from a more canonical domain. 

For Dutch we used a collection of 1,545,871,819 unique Tweets collected between
2010 and 2016, they were collected based on a list of frequent Dutch tokens
which are infrequent in other languages~\cite{42}. For English we collected
Tweets throughout 2016, based on the 100 most frequent words of the Oxford
English
Corpus\footnote{\url{https://en.wikipedia.org/wiki/Most_common_words_in_English}}
, resulting in a dataset of 760,744,676 Tweets.  We used some preprocessing to
reduce the number of types, this leads to smaller models, and thus faster
processing. We replace usernames and urls by $<$USERNAME$>$ and $<$URL$>$
respectively.  As canonical raw data, we used Wikipedia dumps\footnote{cleaned
with WikiExtractor
(\url{http://medialab.di.unipi.it/wiki/Wikipedia_Extractor})} for both Dutch
and English.

\section{Method}
\label{sec:method}
The normalization task  can be split into two sub-tasks:
\begin{itemize}
    \item Candidate generation: generate possible normalization candidates
based on the original word. This step is responsible for an uppperbound on
recall; but care should also be taken to not generate too many candidates, 
since this could complicate the next sub-task.
    \item Candidate ranking: takes the generated candidate list 
from the previous sub-tasks as input, and tries to extract the correct
candidate by ranking the candidates. In our setup we score all the 
candidates, so that a list of top-N candidates can be outputted.
\end{itemize}

Most previous work includes error detection as first step, and only explores
the possibilities for normalization of words detected as anomaly.  However,
we postpone this decision by adding the original word as a candidate.  This
results in a more informed decision whether to normalize or not at ranking
time. We will discuss the methods used for each of the two tasks separately in
the next subsections.

%

\subsection{Candidate Generation}
\label{method:gen}
We use different modules for candidate generation. Each module is focused on a 
different type of anomaly.

\paragraph{Original token}
Because we do not include an error detection step, we need to include the original
token in the candidate list. This module should provide the correct candidate
in 90\% of the cases in our corpora (Table~\ref{tab:corpora}).

\paragraph{Word embeddings} 
We use a word embeddings model trained on the social media domain using the
Tweets described in Section~\ref{data:other}. For each word we find the top 40
closest candidates in the vector space based on the cosine distance. We train a
skip-gram model~\cite{mikolov2013efficient} for 5 iterations with a
vector size of 400 and a window of 1. 

\paragraph{Aspell} 
We use the Aspell spell checker to repair typographical
errors. Aspell uses a combination of character edit distance, and a phonetic 
distance to generate similar looking and similar sounding words.  We will use
the `normal' mode as default, but also experiment with the `bad-spellers' mode,
in which the algorithm allows for candidates with a larger distance to the
original word, resulting in much bigger candidate lists. The effect of this
setting is evaluated in more detail in Section~\ref{sec:aspell}.

\paragraph{Lookup-list}
We generate a list of all replacement pairs occurring in the training data.
When we encounter a word that occurs in this list, every normalization
replacement occurring in the training data is added as candidate.

\paragraph{Word.*} 
As a result of space restrictions and input devices native to this domain,
users often use abbreviated versions of words. To capture this
phenomenon, we include a generation module that simply searches for all words
in the Aspell dictionary which start with the character sequence of our
original word. To avoid large candidate lists, we only activate this module for
words longer than two characters.

\paragraph{Split}
We generate word splits by splitting a word on every possible position
and checking if both resulting words are canonical according to the Aspell
dictionary.  To avoid over-generation, this is only considered for input words
larger than three characters.
\\
\\
To illustrate the effect of these generation modules, the top 3 candidates each
of these modules generate for our example sentence are shown in
Figure~\ref{fig:gen}.  This examples shows that the multiple modules complement
each other rather well, they all handle different types of anomalies.  The
modules are evaluated separately in Section~\ref{sec:evalGen}

\begin{figure}
    \centering
    \tikzset{
system/.style = {draw, fill=white, minimum width=1.5cm, node distance=1.5cm},
rawText/.style  = {text height=1.5ex,text depth=.25ex, align=left,  text width=1.6cm, node distance=1.93cm}, 
combine/.style={draw=gray, opacity=0.7, dashed,thick,inner sep=5pt, minimum width=1.8cm, minimum height=8.4cm}
}

\begin{tikzpicture}[auto, node distance=2cm, scale=0.8, every node/.style={scale=0.88}]
    \node [system] (orig) {orig};
    \node [system, below of=orig] (w2v) {w2v};
    \node [system, below of=w2v] (aspell) {aspell};
    \node [system, below of=aspell] (lookup) {lookup};
    \node [system, below of=lookup] (word.*) {word.*};
    \node [system, below of=word.*] (split) {split};
    
    \node [rawText, right of=orig] (pos10) {\textbf{most}};
    \node [rawText, right of=pos10] (pos20) {\textbf{social}};
    \node [rawText, right of=pos20] (pos30) {pple};
    \node [rawText, right of=pos30] (pos40) {r};
    \node [rawText, right of=pos40, text width=2.2cm, node distance=2.2cm] (pos50) {\textbf{troublesome}};

    \node [rawText, right of=w2v] (pos11) {best/most\\mosy\\MOST};
    \node [rawText, right of=pos11] (pos21) {soical\\Social\\socail};
    \node [rawText, right of=pos21] (pos31) {ppl\\pipo\\ \textbf{people}};
    \node [rawText, right of=pos31] (pos41) {\textbf{are}\\sre\\rnt};
    \node [rawText, right of=pos41, text width=2.2cm, node distance=2.2cm] (pos51) {bothersome\\tricky\\irksome};
    
    \node [rawText, right of=aspell] (pos12) {most\\mist\\moist};
    \node [rawText, right of=pos12] (pos22) {\textbf{social }\\ socially \\ socials};
    \node [rawText, right of=pos22] (pos32) {Pol \\pol \\ Pl};
    \node [rawText, right of=pos32] (pos42) {R \\ r \\ RI};
    \node [rawText, right of=pos42, text width=2.2cm, node distance=2.2cm] (pos52) {\textbf{troublesome} \\trouble some\\trouble-some};
    
    \node [rawText, right of=lookup] (pos13) {\textbf{most}};
    \node [rawText, right of=pos13] (pos23) {\textbf{social}};
    \node [rawText, right of=pos23] (pos33) {};
    \node [rawText, right of=pos33] (pos43) {\textbf{are} \\r\\rest};
    \node [rawText, right of=pos43, text width=2.2cm] (pos53) {};

    \node [rawText, right of=word.*] (pos14) {mostly\\most's};
    \node [rawText, right of=pos14] (pos24) {socially\\social's\\socials};
    \node [rawText, right of=pos24] (pos34) {};
    \node [rawText, right of=pos34] (pos44) {};
    \node [rawText, right of=pos44, text width=2.2cm, node distance=2.2cm] (pos54) {troublesomely};

    \node [rawText, right of=split] (pos15) {mo st};
    \node [rawText, right of=pos15] (pos25) {};
    \node [rawText, right of=pos25] (pos35) {};
    \node [rawText, right of=pos35] (pos45) {};
    \node [rawText, right of=pos45, text width=2.2cm, node distance=2.2cm] (pos55) {trouble some};

    \node [combine] at(2.1,-4.3) (combine1) {};
    \node [combine] at(4.2,-4.3) (combine2) {};
    \node [combine] at(6.3,-4.3) (combine3) {};
    \node [combine] at(8.4,-4.3) (combine4) {};
    \node [combine, minimum width=2.4cm] at(10.83,-4.3) (combine5) {};

\end{tikzpicture}
    \caption{The top 3 generated candidates for each of the generation modules.}
    \label{fig:gen}
\end{figure}


\subsection{Candidate Ranking}
In this section we will first describe the features used for ranking, starting
with the features which originate from the generation step. After this, we
discuss the used classifier.

\paragraph{Original}
A binary feature which indicates if a candidate is the original token. 

\paragraph{Word embeddings}
We use the cosine distance between the candidate and the original word in the
vector space as a feature. Additionally, the rank of the candidate in
the returned list is used as feature.  

\paragraph{Aspell}
Aspell returns a ranked list of correction candidates, we use the rank in this
list as a feature. Additionally, we use the internal calculated distance
between the candidate and the original word; this distance is based on lexical
and phonetical edit distances. The internal edit distance can be obtained from
the Aspell library using C++ function calls. Note that both of these features are
only used for candidates generated by the Aspell module.

\paragraph{Lookup-list}
In our training data we count the occurrences of every correction pair, this
count is used as feature. Note that we also include counts for unchanged pairs
in the training data; this strengthens the decision whether to keep the
original word.

\paragraph{Word.*}
We use a binary feature to indicate if a candidate is generated by this module.

\paragraph{N-grams}
We use two different N-gram models from which we calculate the unigram
probability, the bigram probability with the previous word, and the bigram
probability with the next word. The first N-gram model is trained on the same
Twitter data as the word embeddings, the second N-gram model is based on more
canonical Wikipedia data (Section~\ref{data:other}).

\paragraph{Dictionary lookup}
A binary feature indicating if the candidate can be found in the Aspell
dictionary.

\paragraph{Character order}
We also include a binary feature indicating if the characters of the original
token occur in the same order in the candidate.

\paragraph{Length}
One feature indicates the length of the original word, and one for the length
of the candidate.

\paragraph{ContainsAlpha}
A binary feature indicating whether a token contains any alphabetical
characters; in some annotation guidelines tokens which do not fit this
restriction are kept untouched.
\\
\\
The task of picking the correct candidate can be seen as a binary
classification task; a candidate is either the correct candidate or not.
However, we can not use a binary classifier directly; because we need exactly
one instance for the `correct' class.  Whereas the classifier might classify
multiple or zero candidates per position as correct. Instead, we use the
confidence of the classifier that a candidate belongs to the `correct' class to
rank the candidates. This has the additional advantage that it enables the
system to output lists of top-N candidates for use in a pipeline.

We choose to use a random forest classifier~\cite{breiman2001random} for the
ranking of candidates. We choose this classifier because the problem of
normalization can be divided in multiple normalization actions which behave
differently feature wise, however in our setup they are all classified as the
same class. A random forest classifier makes decisions based on multiple trees,
which might take into account different features. Our hypothesis is that it
builds different types of trees for different normalization actions. More
concretely: if a candidate scores high on the Aspell feature (it has a low edit
distance), this can be an indicator for a specific set of trees to give this
candidate a high score. At the same time the model can still give very high
scores to candidates with low values for the Aspell features.  We use the
implementation of Ranger~\cite{wright2017ranger}, with the default parameters.


\citeasnoun{jin:2015:WNUT} showed that it might have a negative effect on
performance to generate candidates that do not occur in the training data. For
this reason we add an option to MoNoise to filter candidates based on a word
list generated from the training data. Additionally we add an option to filter based
on all words occurring in the training data complemented by the Aspell dictionary; 
these settings are evaluated in more detail in Section~\ref{sec:aspell}.

\section{Evaluation}
\label{sec:evaluation}
In this section, we will evaluate different aspects of the normalization
systems. We evaluate on three benchmarks:
\begin{itemize}
    \item LexNorm1.2: for testing on the LexNorm corpus, we use 2,000 Tweets 
from LiLiu (see Section~\ref{sec:dataNorm}) as training and the other 577 Tweets
as development data.
    \item LexNorm2015: Consisting of 2,950 Tweets for training and
1,967 for testing. We use 950 Tweets from the training set as development data.
    \item GhentNorm: similar to previous work, we split the data in 60\%
training data, and both 20\% development and test data.
\end{itemize}
For all these three datasets we evaluate the different modules of the candidate
generation and the ranking. All evaluation in this section is done with all
words lowercased, because it is in line with previous work and capitalization
is not consistently annotated in the available datasets.

%
%

\subsection{Candidate Generation}
\label{sec:evalGen}
In this section we will first compare each of the modules in isolation. Next,
we test how many unique correct normalization candidates each module 
contributes in an ablation experiment.

\begin{figure}
    \centering
    \begin{minipage}{.45\textwidth}
        \centering
        \includegraphics[width=\linewidth]{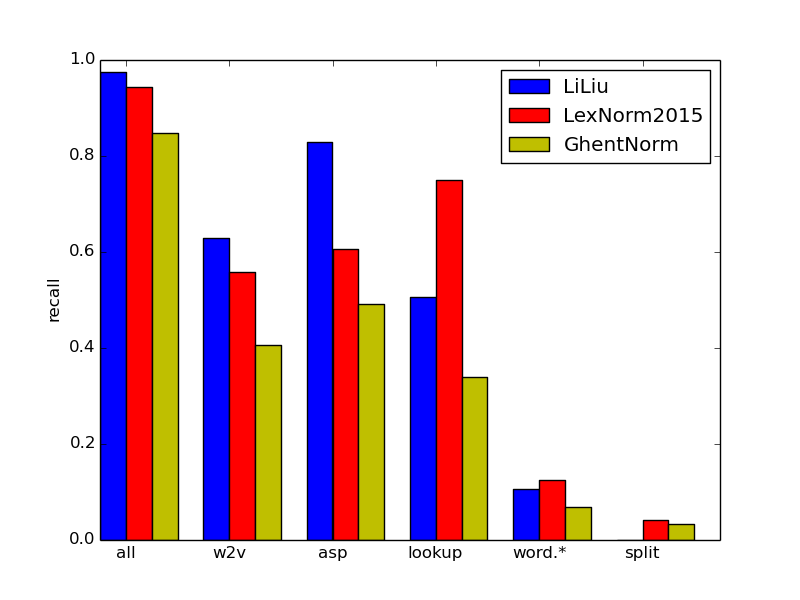}
        \captionof{figure}{Recall of generation modules in isolation on the development corpora.}
        \label{fig:single}
    \end{minipage}%
    \hspace{1cm}
    \begin{minipage}{.45\textwidth}
        \centering
        \includegraphics[width=\linewidth]{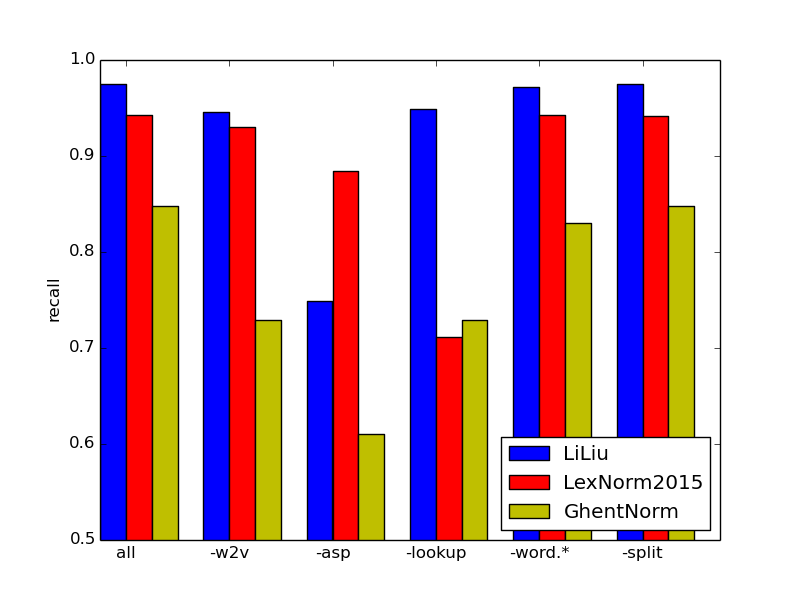}
        \captionof{figure}{Ablation experiments for generation modules on the development corpora.}
        \label{fig:abl}
    \end{minipage}
\end{figure}

\begin{table}
    \centering
    \begin{tabular}{l r}
        Module  & Average candidates \\
        w2v     & 34.13     \\
        Aspell  & 31.29    \\
        lookup  & 0.72      \\
        word.*  & 20.80     \\
        split   & 0.35      \\
    \end{tabular}
    \caption{The number of candidates generated by generation modules, averaged over the different development data sets.} 
    \label{tab:resGen}
\end{table}

The recall of the generation modules in isolation are plotted in
Figure~\ref{fig:single}; the number of candidates each module generates on
average over all datasets is shown in Table~\ref{tab:resGen}. The best performing
modules in isolation are Aspell, word embeddings and the lookup module. The
lookup module performs especially well on the LexNorm2015 corpus. This is due
to a couple of correction pairs which occur very frequently (u, lol, idk, bro).
The word.* module does not perform very well, it over-generates mainly on the
GhentNorm corpus (average of 48 candidates). The split module can only generate 
correct candidates for corpora that contain 1-n word replacements.  For these
corpora, it generates a few correct candidates.

The performances of the ablation experiments are shown in Figure~\ref{fig:abl}.
Similar to the previous experiment, the most important modules are Aspell, word
embeddings and the lookup module. However, the word embeddings
contribute less unique candidates; presumably because it has overlap with both
of the other modules.  The differences between corpora are also similar to the
previous experiment; the lookup list is also generating the most unique correct
candidates for the LexNorm2015 corpus. Furthermore, this graph shows
that word.* still generates some unique candidates, especially for the GhentNorm
corpus. This suggests that this way of abbreviating is more common in Dutch.

\subsection{Candidate Ranking}
\begin{figure}
    \centering
    \begin{minipage}{.45\textwidth}
        \centering
        \includegraphics[width=\linewidth]{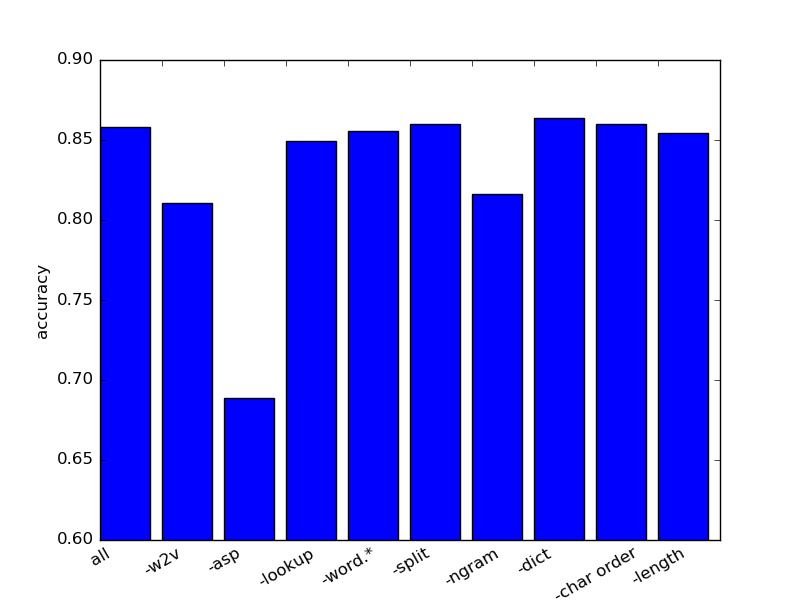}
        \captionof{figure}{Accuracy only on words needing normalization when excluding feature groups on LiLiu development data.}
        \label{fig:chenliAbl}
    \end{minipage}%
    \hspace{1cm}
    \begin{minipage}{.45\textwidth}
        \centering
        \includegraphics[width=\linewidth]{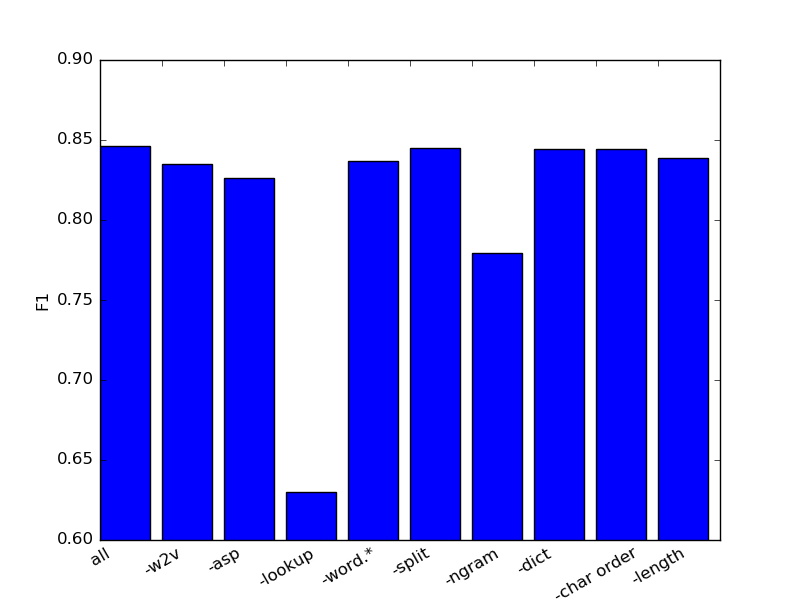}
        \captionof{figure}{F1 scores when excluding feature groups on the LexNorm2015 development data.}
        \label{fig:lexnormAbl}
    \end{minipage}
\end{figure}

Since candidate ranking is the final step, this section discusses the performance
of the whole system. We compare the performance of our system with different 
benchmarks. First we consider the LexNorm corpus, for which most previous work
assumed gold error detection. In order to be able to compare our results, we
will assume the same. Additionally, we test our performance on LexNorm2015,
which was used in the shared task of the 2015 workshop on Noisy User-generated
Text~\cite{baldwin-EtAl:2015:WNUT}.  Here, error detection was included in the
task, hence we will also use automatic error detection by including the
original token as a candidate. 

Figure~\ref{fig:chenliAbl} shows the importance of the different feature groups
in the final model for the LiLiu development set. Aspell is the most important 
feature for this dataset. Except for the split module (which is not included in 
the annotation) all the features contribute to obtaining the highest score.

Figure~\ref{fig:lexnormAbl} shows the F1 scores of the ablation experiments on
the LexNorm2015 corpus. In this setup, the differences are smaller, except for
the lookup module, which generates many unique phrasal abbreviations (lol, idk,
smh).  Perhaps surprisingly, word embeddings show a relatively small effect on
the final performance.  On both datasets, the N-gram module proves to be very
valuable for this task.

\begin{table}
    \centering
    \begin{tabular}{r r r r}
        \#cands & LiLiu & LexNorm2015 & GhentNorm \\
        1       & 95.6  & 97.6        & 98.3\\
        2       & 98.6  & 99.0        & 99.0\\
        3       & 99.1  & 99.2        & 99.1\\
        4       & 99.3  & 99.3        & 99.3\\
        5       & 99.3  & 99.3        & 99.3\\
        ceil.   & 99.6  & 99.4        & 99.8 \\
    \end{tabular}
    \caption{Recall achieved by the top-N candidates for our different development data sets. Note that this is over all words, also words not needing normalization.}
    \label{tab:numCands}
\end{table}

To evaluate the performance of the ranking beyond the top-1 candidate,
Table~\ref{tab:numCands} shows the recall of the top-N candidates on the
different datasets. This table shows that most of the mistakes the classifier
makes are between the first and the second candidate. Manual inspection
revealed that many of these are confusions with the original token; thus the
decision whether normalization is necessary at all. Beyond the second
candidate, there are only a few correct candidates to be found.

\subsection{Additional Experiments}
\begin{figure}
    \centering
    \includegraphics[width=.5\linewidth]{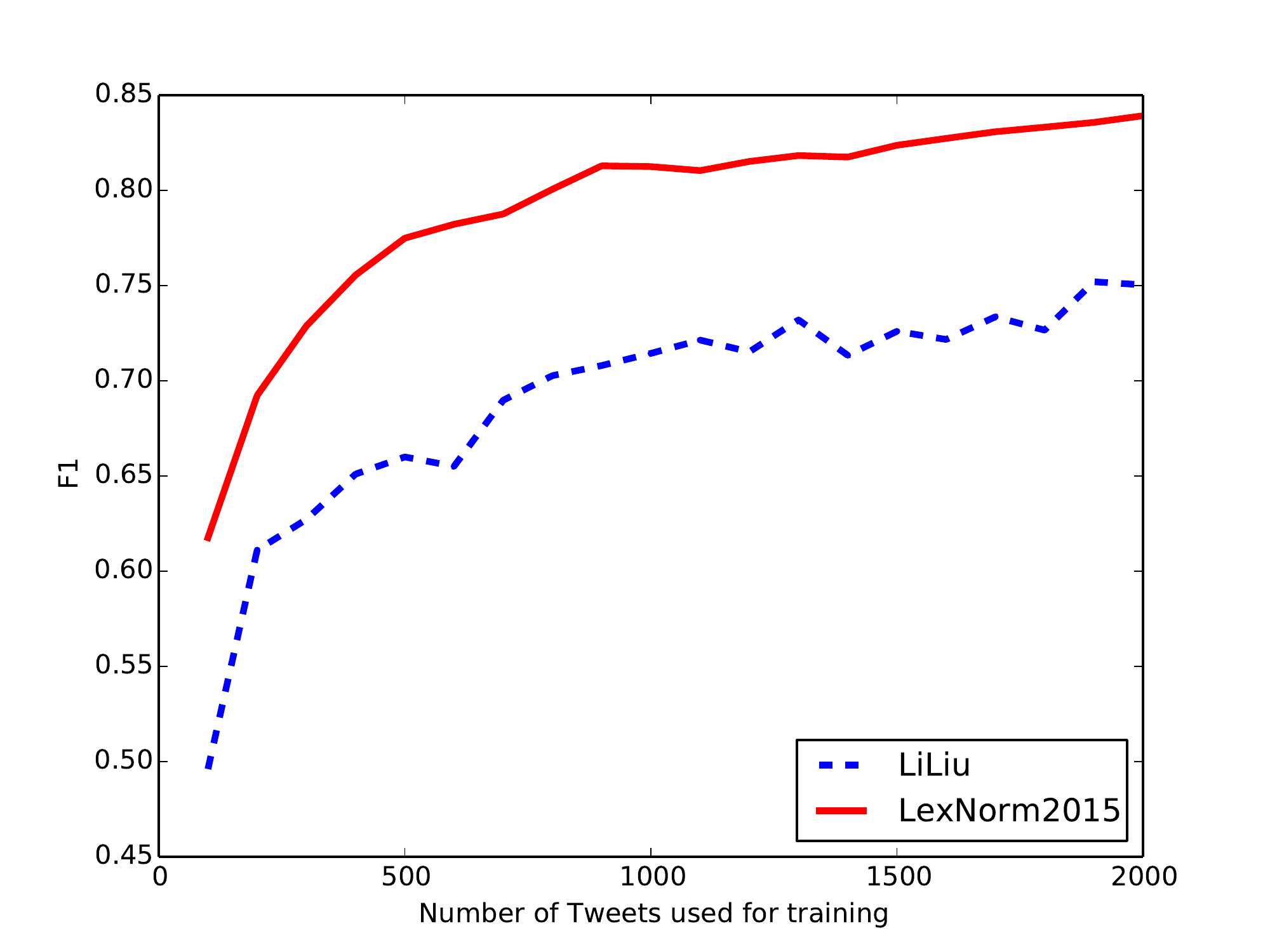}
    \caption{The effect of the size of training data on the LexNorm2015 and LiLiu dataset}
    \label{fig:trainData}
\end{figure}

We test the effect of the size of the training data for the two largest
datasets: LiLiu and LexNorm2015. The results are plotted in
Figure~\ref{fig:trainData}. The higher F1 scores on the LexNorm2015 dataset are
probably due to the common phrasal abbreviations.  Based on these graphs, we can
conclude that a reasonable performance can be achieved by using around 500
Tweets.  However, the performance still improves at a training size of 2,000
Tweets.

\label{sec:aspell}
As explained in Section~\ref{sec:method} we included two options to tune the 
speed-performance ratio. Firstly, we can allow Aspell to generate larger lists 
of candidates by using the `bad-spellers' mode. Secondly, we can filter the generated
candidates, keeping only candidates which occur in the training data or in the 
Aspell dictionary. Table~\ref{tab:aspTime} shows the times the different
combinations of parameters take to train and run on the LexNorm2015 dataset, as
well as the performance and the number of candidates generated. The best
performance is achieved by filtering based on words occurring in the training
data combined with the Aspell dictionary and using the `normal' mode. However,
the `bad-spellers' mode without filtering reaches on par performance and might
be the preferable option, since it can be more robust to data from a different
time period or different domain.

\begin{table}[h]
    \centering
    \begin{tabular}{l l r r r r r}
        Aspell mode  & Filter   & Train time (m:s) & Words/sec. & F1 & Upperbound & Avg. cands. \\
        normal       & -        & 4:56  & 82.4   & 83.4 & 99.5 & 74.2 \\
        normal       & train+asp& 2:35  & 121.2  & 84.2 & 99.2 & 38.7 \\
        normal       & train    & 0:59  & 203.1  & 82.9 & 99.0 & 13.8 \\
        bad-spellers & -        & 28:31 & 23.4   & 84.1 & 99.5 & 306.8 \\
        bad-spellers & train+asp& 19:19 & 31.4   & 84.0 & 99.3 & 208.7 \\
        bad-spellers & train    & 7:12  & 58.8   & 83.3 & 99.1 & 71.9 \\
    \end{tabular}
    \caption{Effect of using different Aspell modes on the LexNorm2015 dataset, using our standard splits (2,000 Tweets train/950 Tweets dev). Train times are in minutes:seconds and averaged over 5 runs; the time needed to load the models is neglected.}
    \label{tab:aspTime}
\end{table}

\subsection{Test Data}
\begin{table}
    \centering
    \begin{tabular} { l l l r r l}
        Corpus      & Prev. source & Eval & Prev.\hspace{.26cm}  & MoNoise & \\
        LexNorm1.2     &\citeasnoun{li2015joint} & acc & 87.58 \tablefootnote{\citeasnoun{li2015joint} use a slightly adapted version of the lexnorm1.2 corpus, MoNoise reaches an accuracy of 88.26 on this data (\url{http://www.hlt.utdallas.edu/~chenli/normalization\_pos/test\_set\_2.txt})} & 87.63    \\ 
        LexNorm2015 &\citeasnoun{jin:2015:WNUT} & F1 & 84.21\hspace{.26cm} &  86.39  \\
        GhentNorm   &\citeasnoun{Schulz:2016:MTN:2906145.2850422} & WER & 3.2\hspace{.26cm}   & 1.7 \\
    \end{tabular}
    \caption{Results on test data compared to the state-of-the-art.}
    \label{tab:test}
\end{table}

The performance of our system on the test data sets is compared with existing
state-of-the-art systems in Table~\ref{tab:test}. Note that different
evaluation metrics are used to be able to compare the results with previous
work. For the LexNorm1.2 corpus we assume gold error detection, in line with
previous work; for more details on the metrics we refer to the original papers.
We use the best settings; meaning the `bad-spellers' mode, no filtering and
including all feature groups.  MoNoise reaches a new state-of-the-art for all
benchmarks. The difference on the LexNorm dataset is rather small; however, our
model is much simpler compared to the ensemble system used
by~\citeasnoun{li2015joint}. The performance gap on the LexNorm2015 dataset is
a bit bigger, showing that MoNoise is also doing well for the error detection
task. Finally, the Word Error Rate (WER) on the Dutch dataset is lower compared
to the previous work.  Note that the evaluation is not directly comparable on
this dataset, since we used different random splits.

Additionally, we report the recall, precision and F1 score for all the
different datasets. We use these evaluation metrics because it allows for a
direct interpretation of the results~\cite{REYNAERT08.477} and it is in line
with the default benchmark of the most recent
dataset~\cite{baldwin-EtAl:2015:WNUT}. We first categorize each word as
follows:
\\
\\
\noindent$TP = $ annotators normalized, systems ranks the correct candidate highest \\
\noindent$FP = $ annotators did not normalize, system normalized \\
\noindent$TN = $ annotators did not normalize, system did not normalize \\
\noindent$FN = $ annotators normalized, but system did not normalize \\

Then we calculate recall, precision and F1 score~\cite{rijsbergen1979v}: 
\\
\\
$Recall = \frac{TP}{TP + FN}$ \\ \\ 
$Precision = \frac{TP}{TP + FP}$\\ \\
$F1 = 2 * \frac{Recall * Precision}{Recall + Precision}$\\

The results are shown in Table~\ref{tab:finalTest}; our system scores better on
precision compared to recall.  Arguably, this is a desirable result, since we
want to avoid over-normalization. For cases where high recall is more
important, we introduce a weight parameter for the ranking of the original
token; in this way we can control the aggressiveness of the model. However,
tuning this weight did not result in a higher F1 score. Our model scores lower
for the GhentNorm corpus, this is partly an effect of having less training data.
However does not explain the complete performance difference (see
Figure~\ref{fig:trainData}), other explaining factors include differences in
language and annotation.

\begin{table}
    \centering
    \begin{tabular}{l r r r}
                    & Recall & Precision & F1 score \\
        LexNorm1.2  & 74.45  & 77.56     & 75.97 \\
        LexNorm2015 & 80.26  & 93.53     & 86.39 \\
        GhentNorm   & 28.81  & 80.95     & 42.50 \\
    \end{tabular}
    \caption{Recall, Precision and F1 score for each of our test sets.}
    \label{tab:finalTest}
\end{table}

\subsection{Extrinsic Evaluation}
To test if this normalization model can be useful in a domain adaptation setup
we used it as a preprocessing step for the Berkeley
parser~\cite{petrov-klein:2007:main}.  We used the resulting best normalization
sequence, but also experimented with using the top-n candidates from the
ranking. We observed an improvement in F1 score of 0.68\% on a Twitter
treebank~\cite{foster2011hardtoparse} when using only the best normalization
sequence and a grammar trained on more canonical data. Whereas, giving the
parser access to more candidates lead to an improvement of 1.26\%.  Note that
the Twitter treebank is less noisy compared to our normalization corpora, which
makes the effects of normalization smaller. For more details on this experiment
we refer to the original paper~\cite{vandergoot:2017:ACL2017}.

Additionally, we tested the performance of the bidirectional LSTM POS tagger
Bilty~\cite{plank-sogaard-goldberg:2016:P16-2}, which we train and test on the
datasets from from~\citeasnoun{li2015joint}. We use the word embeddings model
described in Section~\ref{data:other} to initialize Bilty, as well as character
level embeddings.  This results in a POS tagging model that is already adapted
to the domain to some extent. However, using MoNoise as preprocessing still
leads to an improvement in accuracy from 88.53 to 89.63 and 90.02 to 90.25 on
the different test sets.  More details can be found in the original
paper~\cite{goot:2017:wnut}.

\section{Conclusion}
\label{sec:conclusion}
We have proposed MoNoise; a universal, modular normalization model, which beats
the state-of-the-art on different normalization benchmarks. The model is easily
extendable with new modules, although the existing modules should cover most
cases for the normalization task. MoNoise reaches a new state-of-the-art on
three different benchmarks, proving that it can generalize over different
annotation efforts. A more detailed evaluation showed that traditional spelling
correction complemented with word embeddings combine to provide robust
candidate generation for the normalization task.  If the expansion of common
phrasal abbreviations like `lol' and `lmao' is included in the task, a lookup
list is necessary to obtain competitive performance. For the ranking we can
conclude that a random forest classifier can learn to generalize over the
different normalization actions quite well. Besides the features from the
generation, N-gram features prove to be an important predictor for the
classifier.

Future work includes more exploration concerning multi-word normalizations,
evaluation on different domains and languages, a more in-depth evaluation for
different types of replacements, and the usefulness of using normalization as
preprocessing.  Furthermore, it would be interesting to explore how well an
unsupervised ranking method would compete with the random forest classifier. 

The code of MoNoise is publicly
available\footnote{\url{https://bitbucket.org/robvanderg/monoise} ; all results
reported in this paper can be reproduced with the command
\texttt{./scripts/clin/all.sh}}.

\section*{Acknowledgements}
We would like to thank all our colleagues and the anonymous reviewers
for their valuable feedback and Orph{\'e}e De Clercq for sharing the Dutch dataset. 
This work is part of the Parsing Algorithms for Uncertain Input project, funded by
the Nuance Foundation.

\bibliographystyle{clin} 
\bibliography{bibliography}  

\end{document}